\documentclass[sigconf]{acmart}

\AtBeginDocument{%
  }

\setcopyright{acmlicensed}
\copyrightyear{2025}
\acmYear{2025}
\acmDOI{XXXXXXX.XXXXXXX}

\acmConference[ACM XXXXX]{ACM Conference XX XXXXXXXX, XXXXXXXXXXXXXX, XXX XXXXXXXXXXXX}{XXXX XX--XX,
  2025}{XXXXXX, XXXXXX}

\acmISBN{XXX-X-XXXX-XXXX-X/XX/XX}

\usepackage{xcolor}
\usepackage{hyperref}
\usepackage{subfig}
\usepackage{graphicx}
\usepackage{multirow}

\begin{document}

\title[A Protocol For Relying on Proxy Tasks When Designing CSAI Detection Models]{Minimizing Risk Through Minimizing Model-Data Interaction: A Protocol For Relying on Proxy Tasks When Designing Child Sexual Abuse Imagery Detection Models}

\author{Thamiris Coelho}
\affiliation{%
  \institution{Instituto de Computação, Universidade Estadual de Campinas (UNICAMP)}
  \city{Campinas}
  \country{Brazil}}
  
\author{Leo S. F. Ribeiro}
\affiliation{%
  \institution{Instituto de Ciências Matemáticas e de Computação, Universidade de São Paulo (USP)}
  \city{São Carlos}
  \country{Brazil}
}

\author{João Macedo}
\affiliation{%
 \institution{Departamento de Ciência da Computação, Universidade Federal de Minas Gerais (UFMG)}
 \institution{Departamento de Polícia Federal (DPF)}
 \city{Belo Horizonte}
 \country{Brazil}}

\author{Jefersson A. dos~Santos}
\affiliation{%
  \institution{Department of Computer Science, University of Sheffield}
  \city{Sheffield}
  \country{United Kingdom}}

\author{Sandra~Avila}
\affiliation{%
  \institution{Instituto de Computação, Universidade Estadual de Campinas (UNICAMP)}
  \city{Campinas}
  \country{Brazil}}

\renewcommand{\shortauthors}{Trovato et al.}

\begin{abstract}
  The distribution of child sexual abuse imagery (CSAI) is an ever-growing concern of our modern world; children who suffered from this heinous crime are revictimized, and the growing amount of illegal imagery distributed overwhelms law enforcement agents (LEAs) with the manual labor of categorization. To ease this burden researchers have explored methods for automating data triage and detection of CSAI, but the sensitive nature of the data imposes restricted access and minimal interaction between real data and learning algorithms, avoiding leaks at all costs. In observing how these restrictions have shaped the literature we formalize a definition of ``Proxy Tasks'', i.e., the substitute tasks used for training models for CSAI without making use of CSA data. Under this new terminology we review current literature and present a protocol for making conscious use of Proxy Tasks together with consistent input from LEAs to design better automation in this field. Finally, we apply this protocol to study --- for the first time --- the task of Few-shot Indoor Scene Classification on CSAI, showing a final model that achieves promising results on a real-world CSAI dataset whilst having no weights actually trained on sensitive data.
\end{abstract}



\maketitle

\section{Introduction}
Child sexual abuse is a global problem with severe and lasting consequences for victims. According to the USA's National Center for Missing \& Exploited Children (NCMEC)\footnote{\url{https://www.missingkids.org/cybertiplinedata}}, in 2023, the number of reports of suspected child sexual exploitation was more than 36 million. These reports included more than 100 million image/video files containing child sexual abuse. From 2021 to 2023, the number of child sexual abuse materials increased by more than 20\%, verified by NCMEC's CyberTipline (for the online exploitation of children). In 2023, NCMEC scaled more than 63 thousand reports to law enforcement from urgent cases or the child was in imminent danger, a 140\% increase compared to 2021. 

The proliferation of online platforms has facilitated the spread of child sexual abuse imagery (CSAI), making the development of effective detection tools crucial~\cite{bursztein2019rethinking}. Unfortunately, on a day-to-day basis, it turns out that the volume of material produced is much greater than the capacity for the visual analysis carried out by law enforcement professionals. In this context, the automatic classification of CSAI is~paramount.

Due to legal and ethical barriers --- necessary barriers --- such sensitive data cannot be accessed by anyone besides trained law enforcement agents (LEAs). For this reason, traditional methods use hash comparison to detect CSAI~\cite{Westlake2012ComparingCSAM, UK2021TacklingChildCSAM, DaSilva2012AdaptiveSamplingCSAM, photodna}. While helpful, they are sensitive to minor changes to the file contents, a flaw that allows criminals to keep sharing the same content as long as they are not byte-wise identical to the originals flagged by LEAs~\cite{Schulze2014FeatureFusionCSAM}. Recent efforts have been made to make hash-based comparison more ``perceptual'', but these deep-learning-based comparisons come with other concerns when applied to such sensitive data, as recent methods such as NeuralHash \cite{Apple2021CSAMDetectionTechnical} were shown to be susceptible image manipulations that fool detection and attacks that can surface image details from the produced hashes~\cite{Struppek2022LearningBreakDeep}.

Given the limitations of hash-based methods, attention turned then to the exploration of more robust content-based techniques, from classic bag-of-visual-words matching~\cite{sivic2003video,avila2013pooling} to current deep representation learning~\cite{goodfellow2016deep} techniques. The challenges posed by ethical and legal requirements remain of course, but access to CSAI to evaluate content-based models is possible through partnerships with law enforcement. These partnerships make for more trustworthy studies that were tested on real-world CSAI but do not imply such data is available for training models; there is a severe concern regarding data leak through models given that models can often be reversed to reveal their training data \cite{Carlini2021ExtractingTrainingData} --- to this end, the UK government explicitly prohibits this use of their Child Abuse Image Database even if handled by LEAs \cite{UK2021TacklingChildCSAM}); even models trained with Privacy Preserving techniques are prone to leaking general statistics of datasets \cite{Zhang2021AttributePrivacyFramework}.

The question of minimizing model-data interactions then remains and this unassailable restriction has driven the use of a common pipeline in automatic CSAI recognition: before evaluating on CSAI, researchers often go through the usual CRISP-DM-like \cite{Schroer2021SystematicLiteratureReview} pipeline (of task understanding, data understanding, data preparation, modeling, and evaluation) while working under an accessible, different --- but related to CSAI --- task; In this manuscript we formalize and name these as ``Proxy Tasks'' and further disassemble them as a (i) \textit{related sub-task of CSAI recognition} (e.g., age estimation), which is often combined with (ii) \textit{restrictions common in CSAI} (e.g., few labeled samples). Only after a model is produced under the Proxy Task it is evaluated with CSAI to perform either direct recognition or to use the Proxy Task as a data triage step that accelerates the work of LEAs.

This manuscript then presents the literature on automatic CSAI recognition while specifically framing each study through their use of one or more Proxy Tasks. We then observe gaps in the use of these tasks and present a protocol for developing new systems for automating recognition and triage of CSAI. Our proposal gives weight to consistent input from LEAs, specialists that best know the real data, and explicitly states the role of Proxy Tasks in the design of these new systems and how studies can consider new Proxy Tasks.

We show the effectiveness of this protocol with a case study. Through input from LEAs --- from direct partnerships and the literature --- we identify an under-explored aspect of CSAI with regards to scene composition and locale (a \textit{related sub-task of CSAI recognition}) and study this aspect through the lens of few-shot Learning (given few labeled samples is a \textit{restriction common in CSAI}), introducing for the first time the Proxy Task of few-shot indoor scene classification. Through thorough experimentation under the Proxy Task and final evaluation on CSAI we have found that (i) the state-of-the-art in few-shot learning can be successfully applied to classify indoor scenes in CSAI and that (ii) the features learned under this task are indeed relevant to direct CSAI classification. 

The contributions of this manuscript are then that (1) we introduce the terminology of Proxy Tasks and apply it to current literature; (2) through this new frame we introduce a protocol for choosing new Proxy Tasks and designing new automated systems to aid in CSAI recognition and (3) we demonstrate the effectiveness of this protocol with the introduction of the novel Proxy Task of few-shot indoor scene classification, showing potential in the study of scene features for CSAI and promise in the exploration of Few-shot methods that forgo having learned weights depend on CSAI and contribute to minimizing model-data interactions.

\section{Proxy Tasks and the Literature on CSAI Recognition}

As previously discussed, much of the CSAI recognition literature is dedicated to improving upon the shortcomings of hash matching methods~\cite{de2010nudetective, peersman2016icop, biswas2019boosting,photodna} that are widely used by organizations around the world, including social media moderation practices. The biggest shortcoming of these methods is of course that they are not robust to visual changes and not designed to work with completely new materials being uploaded daily. The major challenge in developing robust and semantic methods for detecting CSAI is how access to real CSAI data is (and should be) restricted. Researchers have then relied on tasks that represent individual challenges but have public data for open evaluation. We call these tasks ``Proxy Tasks''.

\begin{figure}[h]
\centering
\includegraphics[width=1\linewidth]{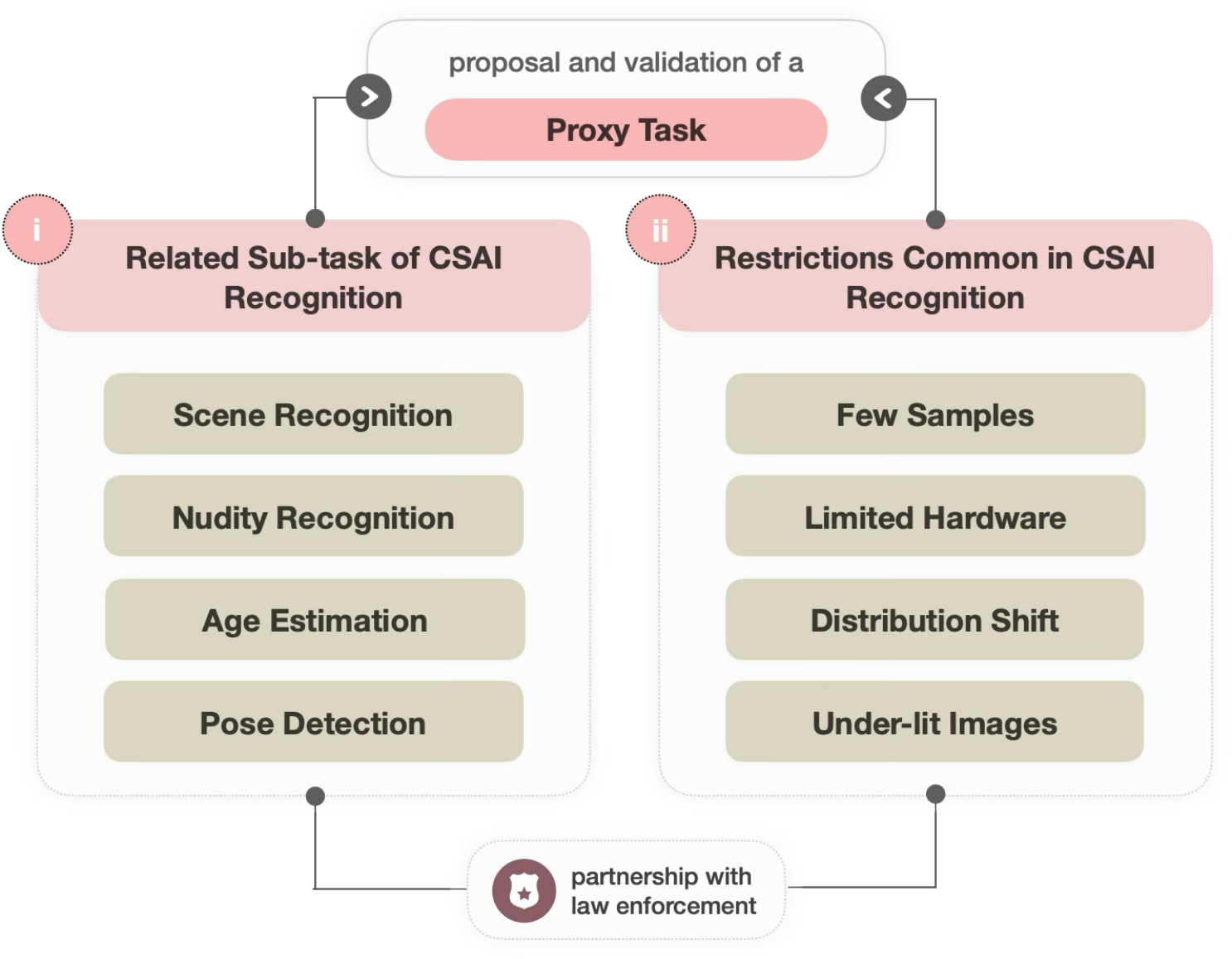}
\caption{The sensitive nature of CSAI requires that most research be done under a related substitute task before being applied to real data. Proxy Tasks combine a related sub-task of CSAI recognition with restrictions common in CSAI data; the figure shows examples of both aspects.}
\Description{A multi-panel diagram showing a set of proxy tasks used for training or evaluation. The layout is grid-like. Labels differentiate between related sub-tasks of CSAI: ``scene recognition'', ``nudity recognition'', ``age estimation'', ``pose detection''; other labels describe restrictions common in CSAI data: ``few samples'', ``limited hardware'', ``distribution shift'', ``under-lit images''.}
\label{fig:proxy-task}
\end{figure}

We formalize these Proxy Tasks as a combination of two aspects, first and foremost, they are a (i) \textit{related sub-task of CSAI recognition}. Identifying what these related sub-tasks are requires thorough communication and partnership with LEAs, the domain experts and stakeholders of CSAI Recognition. As an example, when discussing aspects used by LEAs for manual recognition, Laranjeira da Silva et al.~noted that features like age, gender, face detection, ethnicity, scenes and objects can all offer valuable insights~\cite{laranjeira2022seeing} for a final decision on whether a photograph depicts CSA or not. In addition to a sub-task, we believe it is relevant to consider (ii) \textit{restrictions common in CSAI}; while this is not an aspect that has always been taken into account in the literature, we deem to highlight it given that these restrictions can be a roadblock to implementation and use by LEAs. A non-exhaustive list of such restrictions are: it is not possible to guarantee LEAs will always have the best hardware; model-CSAI interaction should be kept to a minimum; there will always be significant distribution shifts between CSAI and public data. We summarize these two aspects that compose Proxy Tasks in Fig.~\ref{fig:proxy-task}.

We turn our attention now to the literature of robust and semantic CSAI Recognition, identifying within each study the Proxy Task employed. Within the last decade, most research effort have been dedicated to using nudity detection, age estimators and combinations of both. In the studies of \citet{DeCastro2010NudetectiveCSAM, DeCastro2012StatisticalCSAM}, the NuDetective tool was introduced, a nudity detection tool that was effective in detecting nakedness also on CSA images, making for a useful triage model in CSAI recognition. The work of \citet{SaeBae2014TowardsAutomaticCSAM} then combined facial and skin region features to detect nakedness and estimate age, while \citet{Schulze2014FeatureFusionCSAM} also employed nudity detection but combined it with a new Proxy Task, sentiment analysis. This latter study argued that by using nudity detection through skin region features, CSA images were easily confused with legal pornography, images that have very different contexts and image compositions. The use of sentiment analysis however made up for this and showed good potential for detection and explainability. 

While introducing the first deep learning model, \citet{castrillon2018evaluation} relied on the age estimation Proxy Task for their model and fused feature vectors from CNNs with classic ``shallow'' descriptors. The study of \citet{Macedo2018BenchmarkMethodologyChild} made use of a pre-trained Yahoo OpenNSFW model \cite{Mahadeokar2016OpenNSFW} --- a general-purpose pornography detection model --- for nudity detection while also designing an age estimation model that takes face crops as input, again combining these common Proxy Tasks. The study of \citet{vitorino2016wvc,vitorino2018jvci} followed suit, but also showed through a two-tiered fine-tuning pipeline that there is significant benefit in using images from the nudity detection task for finetuning off-the-shelf pre-trained models before use with CSAI. Their hypothesis is that the improvement comes from softening the significant distribution shift between public and sensitive data; this consideration makes it the first study to comment on restrictions common in CSAI recognition, that we argue should be one of the core aspects of a well designed Proxy~Task. 

Used either for triage or for composing a full pipeline, the age estimation task remained popular for applications on CSAI, with multiple studies being dedicated exclusively to differentiating children from adults; The studies of \citet{anda2020deepuage} and \citet{castrillon2018evaluation} have introduced their own datasets, respectively VisAGe and AgeMega, while others have collected images of children from several datasets already available \cite{gangwar2021attm, chaves2020improving}. The practice of using images of children raises concern of course, the UNICEF’s Responsible Data for Children report \cite{Young2019ResponsibleDataUNICEF} strongly recommends against images (or any other related materials) of children to be collected from online sources without proper consent and more specifically without ``principles, practices, and tools that can enable the responsible handling'' of such materials. While this discussion is not the focus of our manuscript, we find it is always appropriate to mention these shortcomings of the research we aim to extend. 

When regarding the other popular Proxy Task of nudity detection, recent studies from \citet{al2020evaluating} and \citet{tabone2021pornographic} have gone for a more specific definition of the task, specifically targeting reproductive organ detection for data triage purposes. The recent study of \citet{valois2024leveraging} introduced the related sub-task of indoor scene classification and studied self-supervised models as they are known to yield general representations that can better adapt to distinct downstream tasks with large distribution shifts. Our case study considers the same related sub-task of indoor scene classification but looks into the restriction of few samples being available for CSAI and studies few-shot methods for indoor scene classification.

In this section then we have shown how the pattern of using Proxy Tasks was present in the literature long before we wrote this manuscript and named them as such. We believe that by highlighting the studies through their Proxy Tasks future work can strive to have results that are more comparable across common tasks even if real CSAI data is not (and should not be) share-able for comparison. We furthermore hope that more Proxy Tasks are proposed, because as \citet{LEE2020301022} pointed out, solutions that achieve the best results --- even when doing manual detection --- are often made with a combination of multiple methods (and therefore, multiple Proxy Tasks). In the next section we present our own protocol for proposing, designing methods for and evaluating Proxy Tasks.  

\section{Protocol for Developing CSAI Detection Solutions}
\label{sec:protocol}

While our protocol in general does not significantly differ from the aforementioned common pipeline of developing models under Proxy Tasks, we use this space to both formalize it and advocate for a few changes in how one should approach CSAI detection. 

At the forefront of these changes, we believe that to further progress with automation when dealing with CSAI we must consider the nascent field of Data-Centric AI. Given the growing presence of AI-driven applications in our daily lives, it is unsurprising that desire has grown for a renewed focus on how these ever-larger models use data and how this data is analyzed, collected and used. A renewed focus then on a data-driven, FAccT-checked (Fairness, Accountability, and Transparency) approach to model design \cite{Polyzotis2021WhatCanDataCentric, Mazumder2022DataPerfBenchmarksDataCentric, Zha2023DatacentricArtificialIntelligence}. When considering CSAI, we see that only through consistent input and partnerships with LEAs we are able to have a better understanding of the available and real-world CSAI data to the interest of designing Proxy Tasks and better benchmarks. 

The first step on our protocol is then to discuss with partner LEAs and look to CSAI literature beyond computer vision approaches to better understand the data, the restrictions and possible Proxy Tasks. Very few studies in the literature match this need for a Data-Centric in CSAI. In the their recent study, \citet{laranjeira2022seeing} designed an analysis framework for child sexual abuse materials (CSAM) data focused on providing valuable insights from the data to researchers without actually giving direct access. Their framework was applied to the Region-based Annotated Child Pornography Dataset (RCPD) \cite{Macedo2018BenchmarkMethodologyChild} and showed how there are many facets to CSAM identification, including nudity and age (as has been explored) but also scene composition and illumination. They have also shown how current deep learning models correlate with real labels regarding object detection and people counting, allowing for automated analysis of these datasets. In the pair of studies from \citet{Kloess2019ChallengesCSAM, Kloess2021Challenges2CSAM}, LEAs were interviewed to highlight the hardest challenges in identifying CSA materials and their findings showed that the environment and objects present in a scene could provide clues as to whether it is inappropriate or not. They also explained that nudity is not an immediate flag for indecency; the posing of subjects and the presence of other images of a victim found together can be better clues of intention. We took these examples together with our own talks with partner LEAs at heart when considering a Proxy Task for our case study presented in Section \ref{sec:case_study}.

\begin{figure*}[h]
\centering
\includegraphics[width=0.95\linewidth]{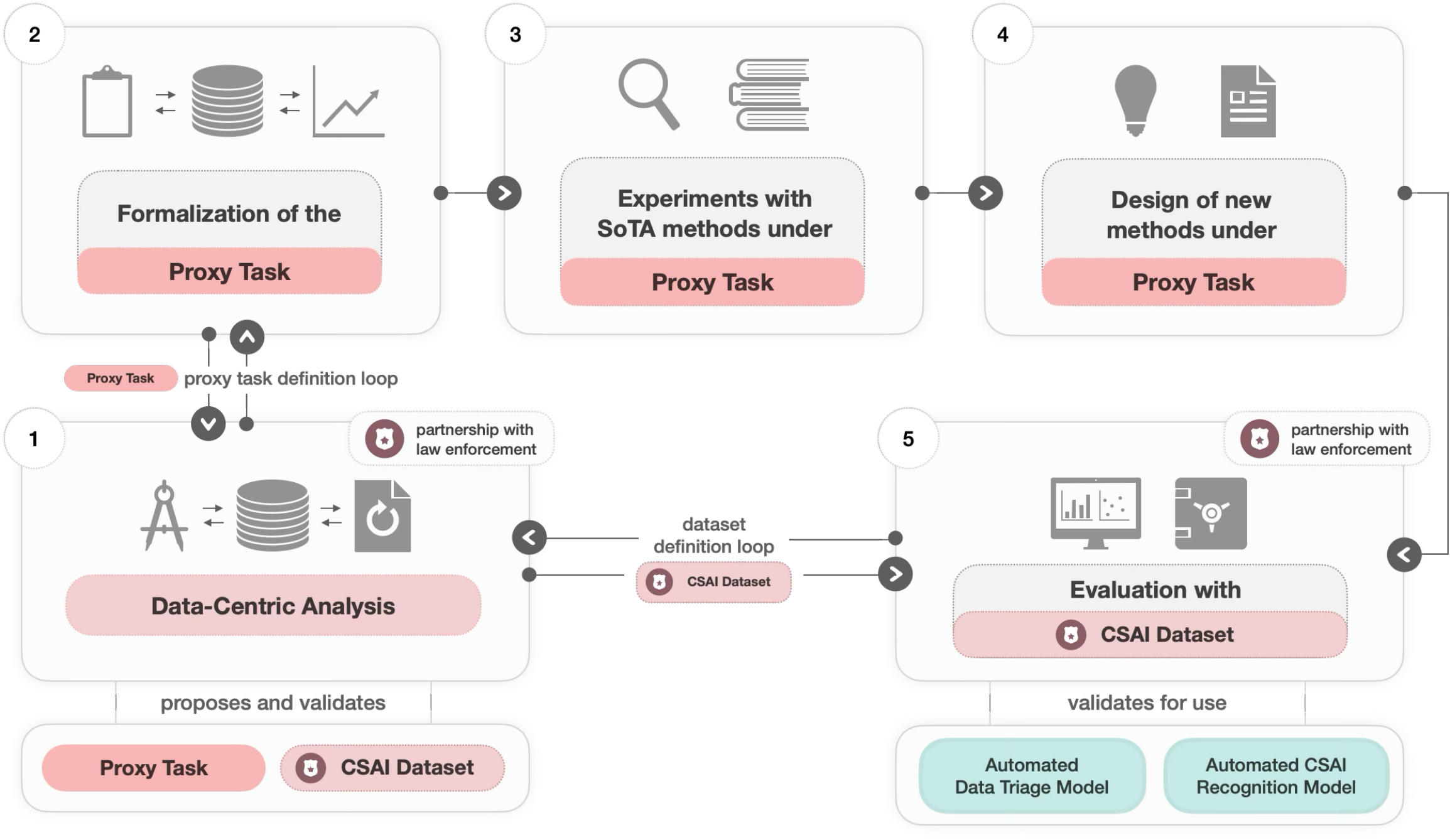}
\caption{Diagram depicting the cycle of steps on our protocol. We believe that Proxy Tasks and CSAI evaluation datasets should be designed with a Data-Centric mindset; Proxy tasks are then used in lieu of real CSAI data to minimize model-CSAI interactions; real CSAI is used only through our partnerships with LEAs to evaluate final models.}
\Description{A block diagram representing the cycle of steps on our protocol. The diagram includes multiple stages connected by arrows, showing the flow of data from input to output. Each block represents one of the steps of the protocol: Data-Centric Analysis, Formalization of the Proxy Task, Experiments with SoTA under Proxy Task, Design of New Methods under Proxy Task and Evaluation with CSAM Dataset. Each block is complemented by an representative icon representing the step (e.g. a pile of discs representing datasets, graphs representing evaluation, light bulbs representing new methods.).}
\label{fig:pipeline}
\end{figure*}

A Data-Centric approach to Proxy Task design is of course only the starting point of our protocol, we better detail all the steps below and organize them in a diagram presented in Fig.~\ref{fig:pipeline}.

\begin{enumerate}
    \item \textbf{Data-Centric Analysis}: We believe that only through consistent input from LEAs --- the stakeholders of these systems --- and thorough consideration of CSAI data we are able to better understand what are the features and their related Proxy Tasks best suited for improved automation of CSAI recognition.  The resulting benchmarks, datasets, and proposed Proxy Tasks from these studies should then drive research independent from these sensitive materials, minimizing model-CSAI interaction.
    \item \textbf{Formalization of the Proxy Task}: With each Proxy Task that has never been attempted before (e.g., our case study, few-shot indoor scene classification), training and evaluation protocols and datasets to benchmark on must be defined.
    \item \textbf{Experiments with SoTA under Proxy Task}: Together with the formal definition of Proxy Tasks, practitioners should define where the current literature is placed in relation to the chosen task and its constraints. A good start is through creating baselines by benchmarking the best models from the related literature (e.g., SoTA models from general few-shot learning literature).
    \item \textbf{Design of New Methods under Proxy Task}: Having established benchmarks and baseline methods, the next step is to to design new methods specifically for the Proxy Task at hand, taking into account both the CSAI-related sub-task and the restrictions considered.
    \item \textbf{Evaluation with CSAM Dataset}: With a new method for the Proxy Task at hand, practitioners should make the necessary adaptations and evaluate the model either/both for automated CSAM recognition and data triage under real-world data. This step is of course done in partnership with LEAs, and its results feed back into the Data-Centric Analysis, closing the loop.
\end{enumerate}

With this protocol at hand we show in our case study how it serves as a guide to the evaluation of a new Proxy Task, few-shot learning for indoor scene classification. We believe that the development and improvement of models through the lens of Proxy Tasks will allow for a future where LEAs have a library of Proxy models to choose from when evaluating a case and such library will make for better data triage, improved ensemble/combined models and better explainability and transparency on (semi-)automated decisions.

\section{Case Study: Few-shot Learning for Indoor Scene Classification}
\label{sec:case_study}

As discussed in Section \ref{sec:protocol}, one of the key intuitions from our initial analysis on Data-Centric studies of CSAI data and our own conversations with LEAs is the importance of features beyond nudity detection, such as the scene composition, placement and position of objects and people in a scene \cite{laranjeira2022seeing, Kloess2019ChallengesCSAM, Kloess2021Challenges2CSAM}. To that end we follow \citet{valois2024leveraging} in basing our Proxy Task on the related sub-task of indoor scene classification. The specificity of only looking at ``indoor'' scenes stems from CSAI being seldom depicted in public spaces and the task of scene classification being a useful way of evaluating aspects of composition, placement and positioning of entities in a scene while also being a task that is well documented and is present in public benchmarks.

In addition to the related sub-task, we believe a good Proxy Task should take into consideration restrictions common when working with CSAI. We know from the literature and our talks with LEAs that large datasets are not available for CSAI (at least not for training models, and definitely not publicly) and it is furthermore unadvisable to incur LEAs with the cost of labeling more data to that end; after all, the mental health crisis in CSA investigation units is a significant concern and a common challenge faced by professionals in this field \cite{strickland2023exploration} and we do not wish to increase that burden. The restriction we consider is then that there are and there will often be very few labeled samples for CSAI. We turn our attention to a particular learning protocol from the general computer vision literature and compose our Proxy Task as \textit{few-shot} indoor scene classification. 

More objectively and following the steps suggested by our protocol, our aim is then \textbf{to evaluate the state-of-the-art in few-shot learning using public data for indoor scene classification}. The best-performing model under this task will then be \textbf{tested with CSAI data through our partnership with LEAs}. 

\subsection{The Literature as it Relates to the Proxy Task}

\textit{Scene classification} aims to classify indoor and outdoor environments. 
While global spacial properties are enough in outdoor environments, both global and local properties are essential in indoor scenes, making the classification a challenge~\cite{patel2020survey}. In those environments, the objects in the scene can be crucial to the classification~\cite{qiu2021scene}. 
 
Classic supervised methods for scene classification have achieved notable success on benchmarks. FOSNet~\cite{seong2020fosnet} achieved an accuracy of 90.3\% for MIT Indoor~\cite{quattoni2009recognizing} and 
77.3\% for \text{SUN-397}~\cite{xiao2010sun}; while the accuracy for Places365, a more robust dataset, drops significantly to 60.1\%. OmniVec2~\cite{Srivastava_2024_CVPR}, the state-of-the-art for Places365, achieved 65.1\%. These methods are however designed to be trained with large datasets and we turn our attention instead to methods from the general few-shot learning literature.

\textit{Few-shot learning (FSL)} is a machine learning method that learns from a few labeled examples, unlike traditional methods that require large annotated datasets. It is particularly useful for real-world scenarios where obtaining extensive labeled data is unfeasible. Most FSL methods learn how to transfer knowledge from a large dataset related to the target task to a small labeled dataset with samples of interest~\cite{hu2022pmf}.

Many studies in the field adopt a meta-learning paradigm, where the model learns to learn~\cite{sung2018relation,snell2017prototypical,vinyals2016matching,finn2017maml}. For that, a model learns from the large dataset (meta-training), mimicking the few-shot task; then, the learned model is applied to the target task dataset (meta-testing). The learned model can be used as model initializer~\cite{finn2017maml}, optimizer~\cite{ravi2016optimization}, or metric learner~\cite{snell2017prototypical}. 

In this case study, we focus on metric learning approaches, which aim to learn how to compare samples. If a model knows how to measure the similarity between samples, it can also classify unseen samples by comparing them~\cite{koch2015siamese}. More specifically, we employ embedding learning~\cite{snell2017prototypical, sung2018relation}, a technique under the metric learning umbrella. The idea is to learn how to generate the best embedding function for the samples so that same-class sample embeddings are closer to embeddings of samples from different classes. The core advantage of this approach is that once the embedding function is learned the weights are not updated on the target task and classification is performed only through comparing the learned representations; for our use case that means that no weights are trained or learned from sensitive data, minimizing model-data interaction. 

\subsection{Few-shot Learning Problem Definition}

In FSL, the goal is to classify unseen samples using only a few labeled data. Such a task is referred to as $N$-way $K$-shot classification, where $N$ is the number of classes considered and $K$ is the number of labeled samples per class. It is common for FSL methods to consider a transfer learning approach, where a model is first pre-trained on a base set $D_{train}$ and then the FSL evaluation considered under a separate unseen set $D_{test}$, where the classes from each set are disjointed: $C_{train} \cap C_{test} = \emptyset$. 

We opted for methods that follow this approach specifically so that knowledge accumulated from public data of scene imagery can later be employed within the domain of CSAI. In further detail, this pre-training stage follows the \textit{episodic meta-learning} concept introduced by Vinyals et al.~\cite{vinyals2016matching}; for each training step, an episode is built by simulating one \text{$N$-way} $K$-shot task. The episode $E$ is composed of a support set $S_{sup} = \{S_{sup}^{nk} = \{x_{sup}^{nk}, y_{sup}^{nk}\}~|~n = 1, ..., N; k = 1, ..., K\}$ and a query set $S_{qry} = \{S_{qry}^{n} = \{x_{qry}^{n}, y_{qry}^{n}\}~|~n = 1, ..., N\}$, where the former represents a small training set and the latter a test set in the simulated task. \textit{Episodic meta-learning} then is to learn --- often through end-to-end gradient descent --- to improve performance on these episodes. A matching protocol, \textit{episodic meta-testing}, is used for evaluating methods; the difference is that within meta-training episodes are sampled from the base set $D_{train}$ and within meta-testing from the unseen set $D_{test}$; because of our specific choice of using embedding learning methods, no gradient steps are taken during \textit{episodic meta-testing} and learning is done through comparing sample representations (i.e., as in a nearest neighbors classifier).

\subsection{Comparative Analysis}

\subsubsection{Methodology}

We chose eleven FSL methods from the literature for comparison. Most of these comply with the Embedding Learning paradigm for FSL; differences can be found in model architecture and specific loss designs. Specific hyperparameters were replicated from the original studies.

\paragraph{Pre-training Datasets}  Methods are pre-trained on \textit{mini}ImageNet, with P>M>F~\cite{hu2022pmf} and CPEA~\cite{Hao2023ClassAwarePatchEmbedding} being~the exceptions  presenting models trained with \text{ImageNet-1K~\cite{deng2009imagenet}.}

\paragraph{Fine-tuning Dataset} This is the first study to address FSL for scene classification. Testing is to be done with the Places8 dataset, proposed by Valois et al.~\cite{valois2024leveraging} to represent indoor scenes common in CSAI. It is also sensible to have a Base set within the domain of scenes. Following the design of \textit{mini}ImageNet, we present \textit{Places600}, a subset of Places365~\cite{zhou2017places} designed to have 600 samples per class, excluding classes with insufficient samples, and not to have class overlap with Places8. We ended up with a dataset that contains 268 classes. \textit{Places600} is used to fine-tune methods for the domain of scenes. The design of \textit{Places600} and the choice of Places8 as a testing ground are part of what we defined as the (2) Formalization of the Proxy Task step within our protocol. 

\paragraph{Training and Evaluation Details} Unless specified otherwise by the original authors, models were trained for 100~epochs composed of 2,000 episodes each. For evaluation, 10,000~episodes are randomly sampled from Places8 with the 8-way 5-shot setup, allowing all 8 classes to participate in all tests. Each evaluation episode contains 15 test queries per class. A single  NVIDIA Tesla T4 was used in all experiments, and the training time for each experiment was an average of less than two days. All code and data (\textit{Places600)} are available at \url{https://github.com/thamycoelho}.

\subsubsection{Results}

The comparison of state-of-the-art FSL methods and selection of a best performing model on the selected indoor scenes benchmark comprise steps (3) and (4) of our protocol. We report the comparison result in Table~\ref{tab:fsl_evaluation}. For a fair comparison, these results all used \textit{mini}ImageNet for pre-training. We can see that the model based on PMF, fine-tuned on \textit{Places600}, was the best performing. The core contribution of the authors of PMF \cite{hu2022pmf} was showing that an updated architecture combined with larger-scale pre-training surpassed other approaches in FSL. With their proposed model performing the best, we follow their experimental design and compare PMF and second-best method CPEA using different backbone networks and the full ImageNet-1K dataset for pre-training.

The result of this comparison can be found in Table~\ref{tab:fsl_evaluation_in1k}. PMF continues to perform better. More importantly, these results surpass the ones that used \textit{mini}ImageNet by a large margin, corroborating Hu et al.~\cite{hu2022pmf}'s findings. We also show how PMF performs with a ResNet-50 backbone and confirm that the newer Transformer-based ViT is the better choice. Our final model is then a PMF model, pre-trained on \text{ImageNet-1K} with the ViT Small backbone; with this model at hand, we have further analyzed its hyperparameters and found that through using stepped learning rate decay every 10 epochs and a temperature of 0.07, an accuracy of 72.50\% is achieved.

\begin{table}[t]
\centering
\caption{Results of few-shot methods on Places8 validation set. Pre-trained and fine-tuned top-1 accuracy results are reported. Best result presented in bold.}
\label{tab:fsl_evaluation}
\begin{tabular}{lcccc}
\toprule
Method & Backbone & Pre-trained & Fine-tuned \\ \midrule
ProtoNet~\cite{snell2017prototypical} & ResNet-12 & 37.48 {\scriptsize ± 0.10} & 43.13 {\scriptsize ± 0.10} \\
MAML~\cite{finn2017maml} & Conv-64 & 34.42 {\scriptsize ± 0.09} & 36.42 {\scriptsize ± 0.10} \\
RelationNet~\cite{sung2018relation} & ResNet-12 & 30.49 {\scriptsize ± 0.09} & 38.69 {\scriptsize ± 0.10} \\
Baseline++~\cite{closerlokfewshot} & ResNet-18 & -- & 38.36 {\scriptsize ± 0.09} \\
LEO~\cite{rusu2018leo} & ResNet-18 & 31.09 {\scriptsize ± 0.09} & 32.66 {\scriptsize ± 0.91} \\
FEAT~\cite{ye2020feat} & ResNet-18 & 45.43 {\scriptsize ± 0.09} & 45.34 {\scriptsize ± 0.10} \\
CTransf.~\cite{doersch2020crosstransformers} & ResNet-34 & 46.67 {\scriptsize ± 0.10} & 45.07 {\scriptsize ± 0.22} \\
PMF~\cite{hu2022pmf} & ViT Small & 45.49 {\scriptsize ± 0.10} & \textbf{52.86} {\scriptsize ± 0.10} \\
HCTransf.\cite{he2022hctransformers} & ViT Small & 44.39 {\scriptsize ± 0.10} & 44.03 {\scriptsize ± 0.10} \\
SSFormers~\cite{ssformers} & ResNet-12 & 41.20 {\scriptsize ± 0.11} & 46.27 {\scriptsize ± 0.12} \\
CPEA~\cite{Hao2023ClassAwarePatchEmbedding} & ViT Small & 49.25 {\scriptsize ± 0.30} & 51.65 {\scriptsize ± 0.30} \\ \bottomrule
\end{tabular}
\end{table}

\begin{table}[]
\centering
\caption{Results of best-performing few-shot methods, pre-trained on ImageNet-1K to show the impact of large-scale data for pre-training. Top-1 accuracy on the Places8 validation set~shown.}
\label{tab:fsl_evaluation_in1k}
\begin{tabular}{lccc}
\toprule
Method & Backbone & Pre-trained & Fine-tuned \\ \midrule
\multirow{2}{*}{PMF~\cite{hu2022pmf}} & ViT Small & 69.76 {\scriptsize ± 0.09} & \textbf{71.86} {\scriptsize ± 0.10} \\
 & ResNet-50 & 59.63 {\scriptsize ± 0.10} & 61.29 {\scriptsize ± 0.10} \\
CPEA~\cite{Hao2023ClassAwarePatchEmbedding} & ViT Small & 64.74 {\scriptsize ± 0.30} & 67.79 {\scriptsize ± 0.30} \\ \bottomrule
\end{tabular}
\end{table}

\begin{figure}[t!]
\centering
\includegraphics[width=\linewidth]{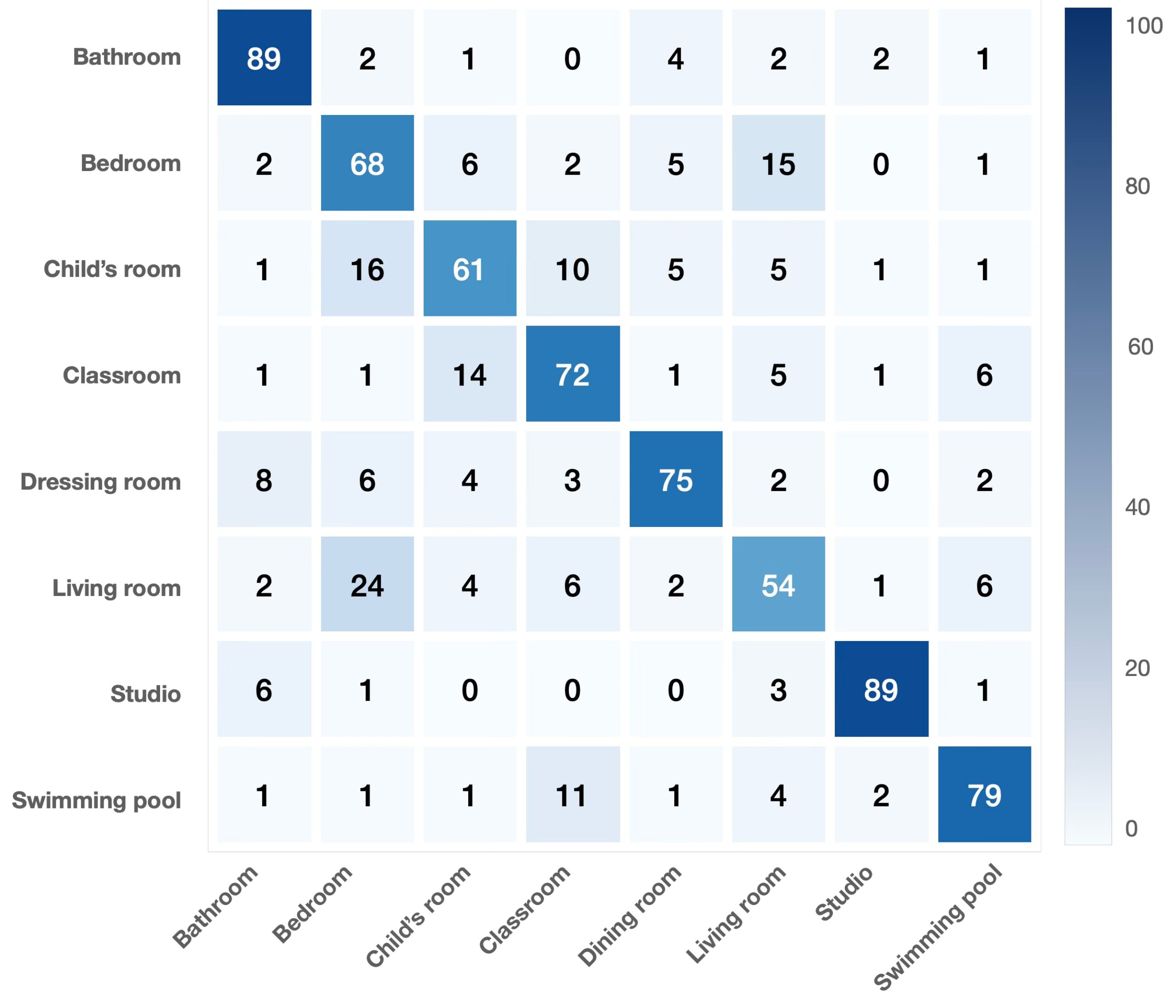}
\caption{Confusion matrix with the results of the model on the Places8 test set. The confusion matrix presents the accuracy~(\%) of the predicted classes following the FSL evaluation protocol. Ground truth labels are on the vertical axis; predicted labels on the horizontal axis.}
\Description{A confusion matrix grid visualization. The grid shows ground truth labels on the vertical axis and predicted labels on the horizontal axis. Each cell is shaded to indicate accuracy or frequency, with darker shades representing higher values. The matrix highlights correct predictions along the diagonal and errors off-diagonal.}
\label{fig:cm_places_final} 
\end{figure}

\begin{figure*}[t!]
\centering
\includegraphics[width=0.65\linewidth,clip, trim={0.0cm 0.0cm 0.0cm 3.5cm}]{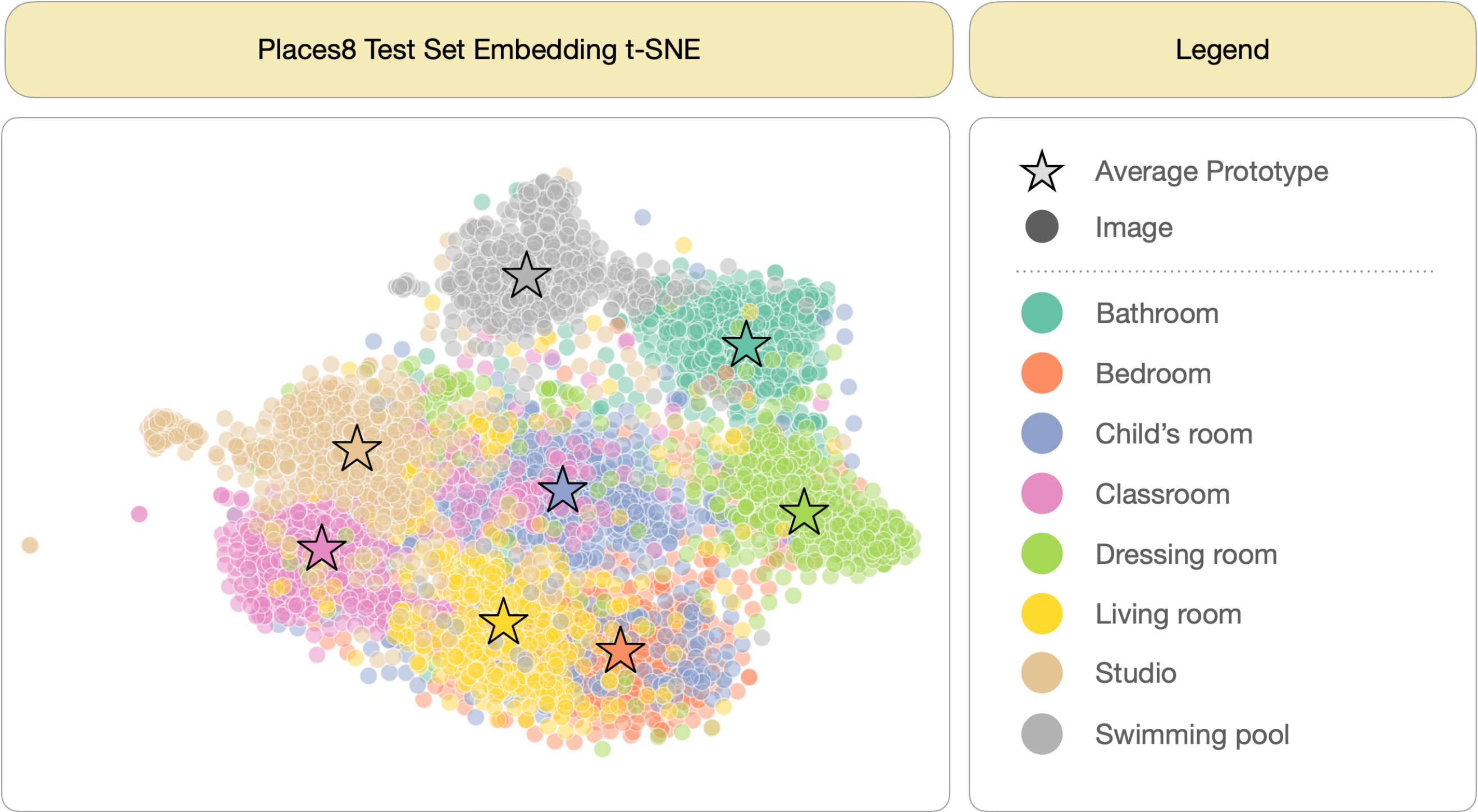}
\caption{t-SNE on two dimensions for the Places8 test set. Many children's room samples (blue dots) are closer to the bedroom prototype (orange star) than to the child's room (blue star). We observe the same phenomenon between classroom samples (pink dots) and child's room samples (blue dots).}
\Description{A 2D scatter plot generated using t-SNE, visualizing high-dimensional feature embeddings. The plot contains multiple colored clusters, each representing different categories or classes. Points that are close together in the plot indicate similar representations in the high-dimensional space. Average vectors for each class are represented by a star symbol within the visualization.}
\label{fig:tsne_places_test} 
\end{figure*}

With our best model at hand, we can move our analysis to the test set of Places8. The model achieved an average accuracy of 73.50~±~0.09\%. Through the confusion matrix (Fig.~\ref{fig:cm_places_final}),
we can observe that \textit{bedroom}, \textit{living room} and \textit{child's room} were often misclassified between them; there is also confusion between \textit{child's room} against \textit{bedroom} and \textit{classroom}. To better observe these missteps, we plot the embeddings from our model using t-SNE~\cite{vd2008tsne}; Fig.~\ref{fig:tsne_places_test} shows feature vectors for these classes have their clusters but that they indeed overlap with each other.

The standard FSL evaluation protocol makes use of test samples to compose episodes. To push the boundaries of our model beyond FSL, we design a custom \textit{Comparable Protocol}. To avoid what would be characterized as a ``leak'' in traditional classification, our protocol samples the support set (5-shot, 5 samples per class) from the validation set of Places8 and uses the entire test set (40,534 samples) for evaluation for each episode. When we follow the standard FSL evaluation protocol, the query set in each epoch is balanced, but we can not assume that for the \textit{comparable protocol}. Importantly, this protocol does not relinquish the claim to being a FSL protocol; we after all use only a few samples (5) in the support set. The difference is that the support and the query sets can have samples from different~domains and, more importantly, not using test samples to compose support sets means the results obtained with this protocol are \textit{comparable} against other traditional machine learning uses of the benchmark.

Under the \textit{comparable protocol}, we achieve an accuracy of 69.25~± 0.05\%.~This is a competent result; \text{Valois et al.} reported 71.6\% with their self-supervised pipeline while actually fine-tuning weights on the entirety of the large Places8 training set; our model, on the other hand, used only 5 samples per Places8 class and did not employ fine-tuning, only comparisons among sample representations.

\subsection{Evaluation on Child Sexual Abuse Imagery}

We conducted the final tests in collaboration with a Brazilian Federal Police agent, per step (5) on our protocol: Evaluation with CSAI Dataset. The agent annotated 374 randomly chosen samples with the indoor classes to evaluate the model on indoor scenes. This set was sampled from an internal collection of images (4,592) categorized into: \textit{CSAI}, \textit{suspected CSAI}, \textit{drawing}, \textit{people}, \textit{porn}, and \textit{other}. As the samples were randomly selected, they only contained 6 out of 8 classes from Places8: \textit{bathroom}, \textit{bedroom}, \textit{child's room}, \textit{living room}, \textit{studio}, and \textit{swimming pool}.

\begin{figure*}[t!]
    \centering
        \subfloat[Few-shot protocol\label{fig:cm_csam_indoor_fsl}]{\includegraphics[width=\columnwidth]
        {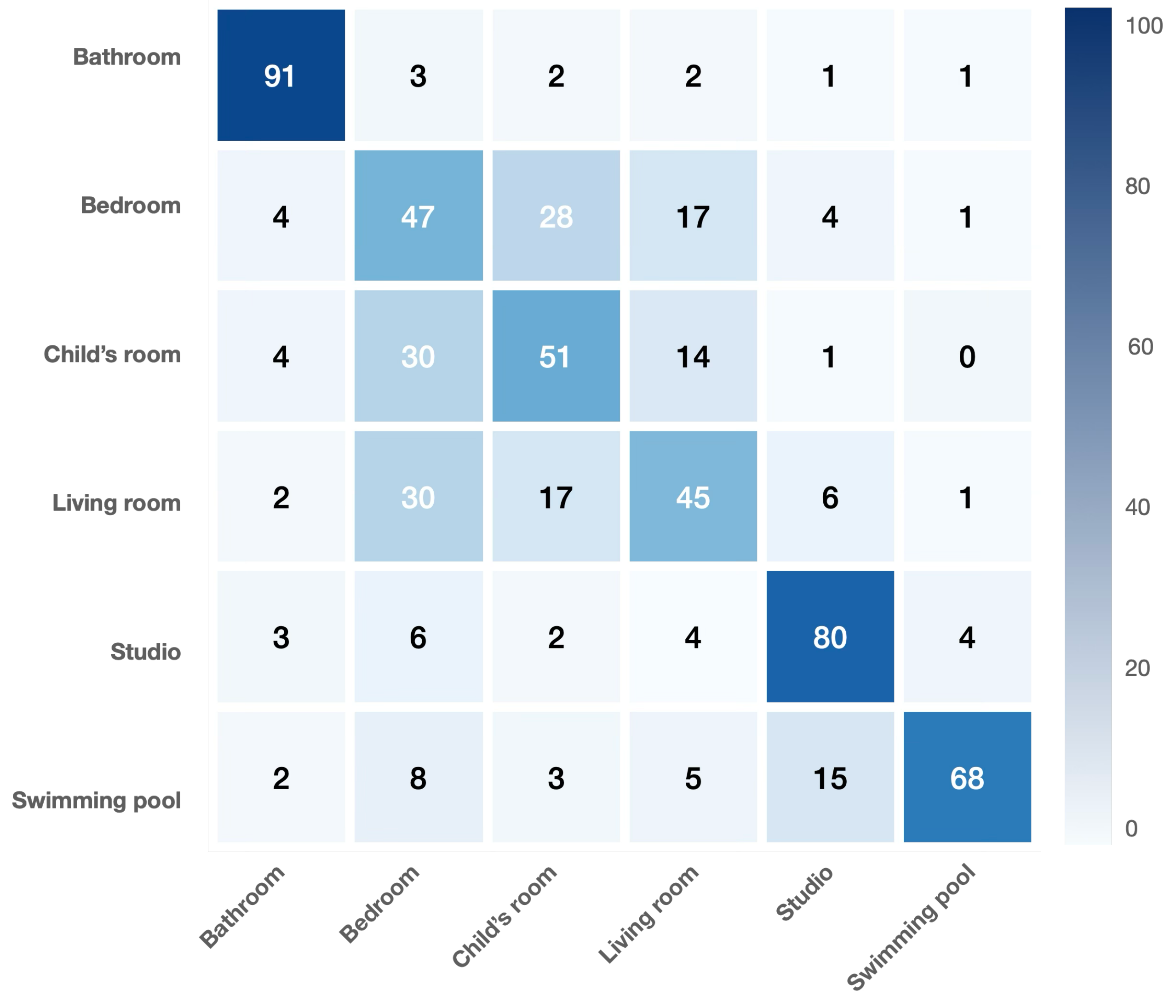}}
        \subfloat[Comparable protocol \label{fig:cm_csam_indoor_general}]{\includegraphics[width=\columnwidth]
        {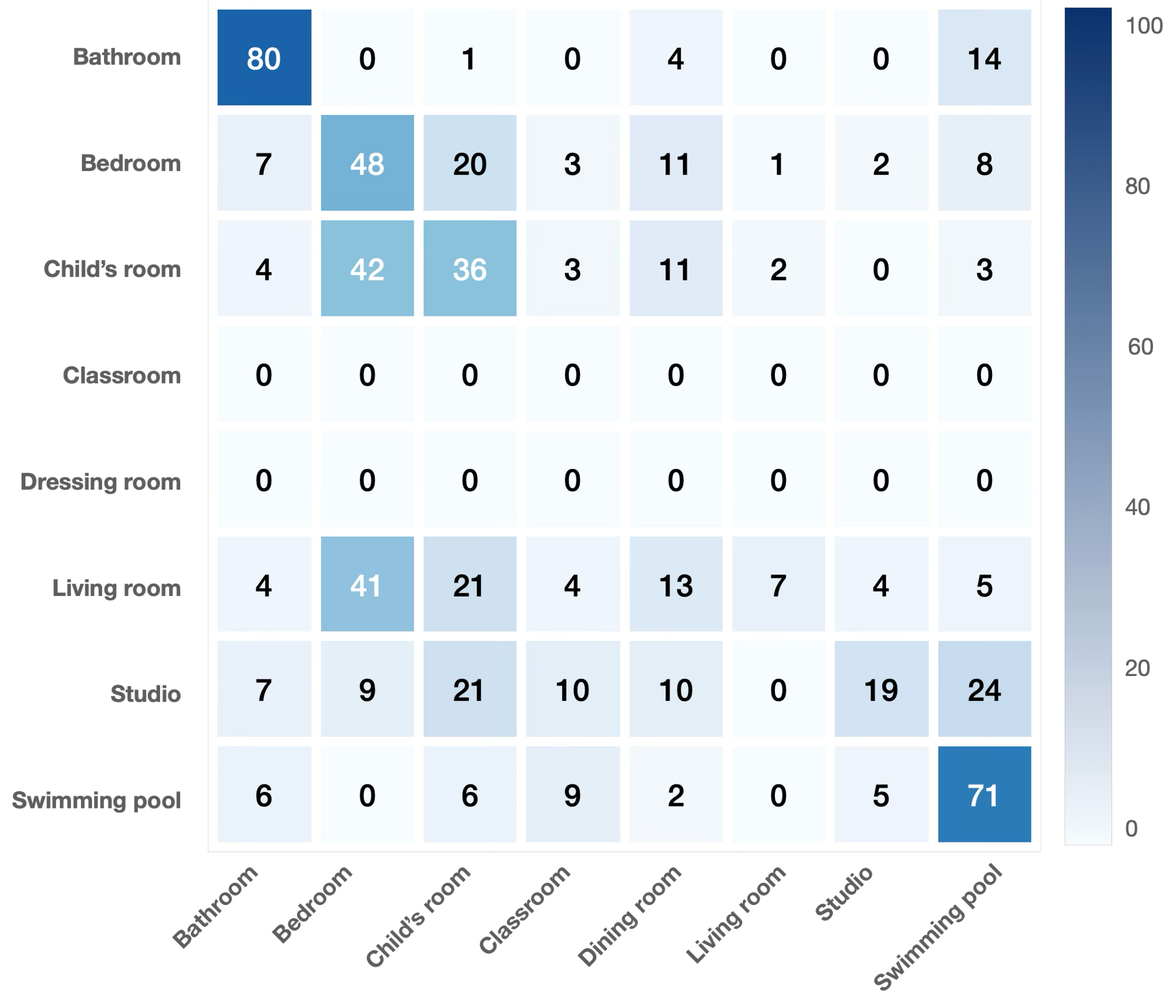}}
        \caption{Confusion matrix with the model's results on the CSAI samples classified into indoor. Values reported are the accuracy~(\%) for prediction per class. Ground truth labels are on the vertical axis; predicted labels on the horizontal axis.} 
        \Description{A confusion matrix grid visualization. The grid shows ground truth labels on the vertical axis and predicted labels on the horizontal axis. Each cell is shaded to indicate accuracy or frequency, with darker shades representing higher values. The matrix highlights correct predictions along the diagonal and errors off-diagonal.}
    \label{fig:cm_csam_indoor}
\end{figure*}

We followed both the traditional FSL evaluation and our \textit{comparable protocol}. In both, we evaluated the model for 10,000 episodes. For the few-shot evaluation protocol, instead of selecting 15 samples per class to compose the query set, due to the limited number of annotated samples, all remaining samples not in the support set were used to compose the query set. Fig.~\ref{fig:cm_csam_indoor} shows the confusion matrix for both experiments.

For the few-shot protocol, the average accuracy from the episodes with 95\% confidence is 63.38 ± 0.09\%. From the confusion matrix, the major confusion occurs among \textit{bedroom}, \textit{child's room}, and \textit{living room}, a similar behavior observed in the evaluation on Places8. The model was good at classifying \textit{bathroom} and \textit{studio}. Given those results, we believe the few-shot proposed model is generalized for the CSAI environment from only a few samples.

For the \textit{comparable protocol}, the model could classify well \textit{swimming pool} and \textit{bathroom}; there was also a considerable confusion between \textit{bedroom} and \textit{child's room}, which was expected given the results on Places8. The model could not classify \textit{living room}, classifying most of the samples as \textit{bedroom} and \textit{child's room}. For the samples from \textit{studio}, the model is also confused with \textit{swimming pool} and \textit{child's room}. A similar behavior was observed by Valois et al.~\cite{valois2024leveraging}, who obtained a balanced accuracy of 36.7\%. Our model achieved a balanced accuracy of 43.43 ± 0.09\% with 95\% confidence.

Although the results on the CSAI indoor dataset were lower than the results obtained on Places8, indicating a domain shift, the model still performed better than the self-supervised model proposed by Valois et al.~\cite{valois2024leveraging}. This suggests that the proposed model is less specialized for Places8 and more capable of generalization. As for the difference between this result with the \textit{comparable protocol} and the FSL protocol presented before, we highlight that on the latter each episode's support set is composed of images from the CSAI indoor collection while the former uses public images from Places8 as support; this distinction is further evidence of both the domain shift between Places8 and CSAI and of our model's capability for generalization with few samples.

\begin{figure}[t!]
\centering
\includegraphics[width=\linewidth]{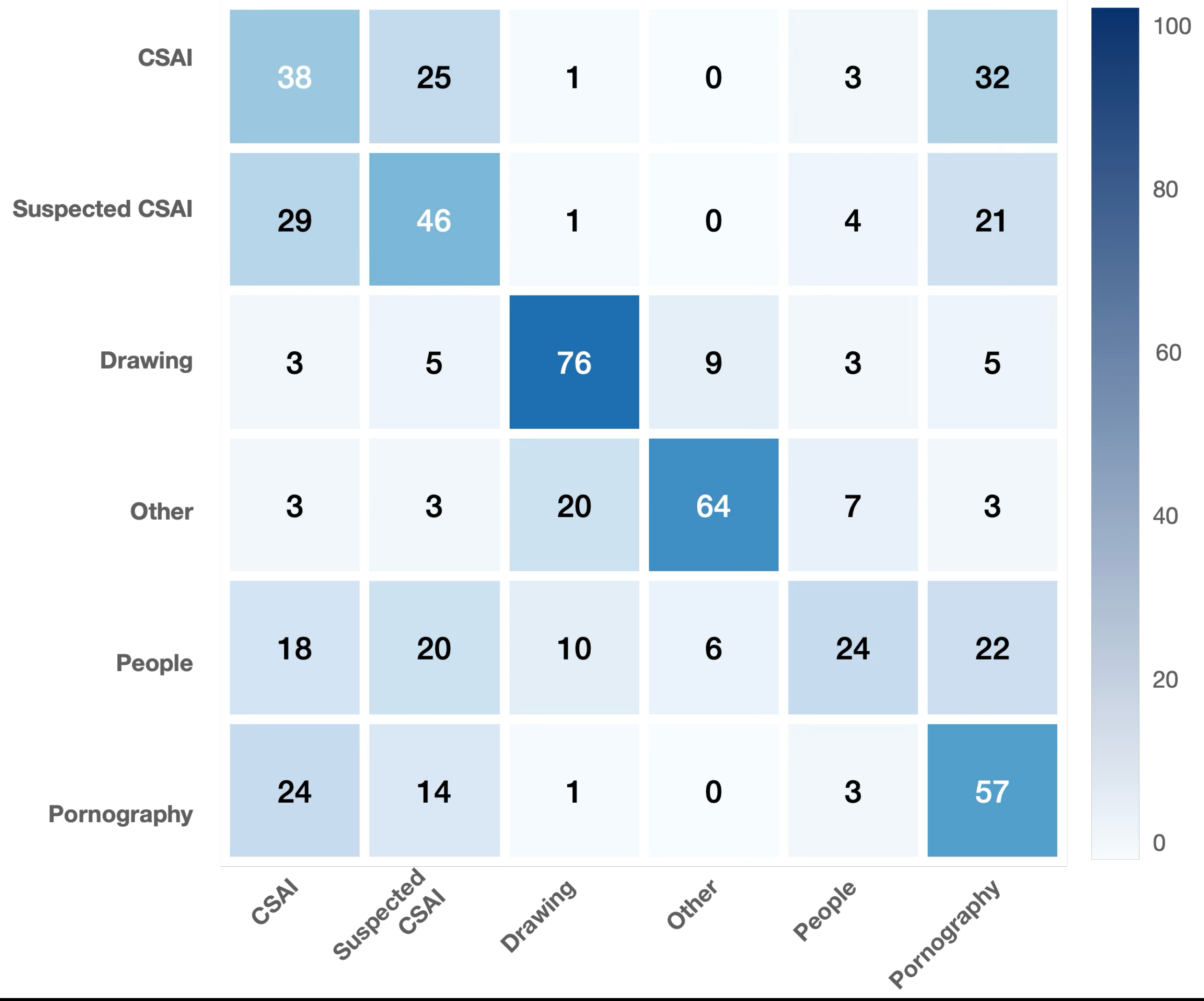}
\caption{Confusion matrix with the model's results on the samples with CSAI annotation. Values reported are the accuracy~(\%) for prediction per class. Ground truth labels are on the vertical axis; predicted labels on the horizontal axis.}
\Description{A confusion matrix grid visualization. The grid shows ground truth labels on the vertical axis and predicted labels on the horizontal axis. Each cell is shaded to indicate accuracy or frequency, with darker shades representing higher values. The matrix highlights correct predictions along the diagonal and errors off-diagonal.}
\label{fig:cm_csam} 
\end{figure}

To understand the importance of the features of indoor scenes to the CSAI recognition itself, since the model is trained to compare samples, we evaluated it on the 4,592~internal collection of images; none of the classes from this collection are present in Places8. Therefore, we only performed the FSL evaluation protocol. Fig.~\ref{fig:cm_csam} shows the confusion matrix of the experiment.

The model achieved an average accuracy through all the episodes for the CSAI dataset with 95\% confidence of \text{50.82 ± 0.11\%}. From the confusion matrix, we can observe confusion among \textit{CSAI}, \textit{suspected CSAI}, and \textit{porn}. It makes sense, given that those samples are probably visually similar, making it difficult even for police agents to differentiate those classes. However, the model effectively classified \textit{drawing}, possibly due to their distinct domain from the training data. There was a slight confusion between \textit{other} and \textit{drawing}, probably because the class \textit{other} contains images of documents, banknotes, and handwritten documents. Hence, the model probably understood some drawings as documents and vice versa. The model struggled to classify \textit{people}, often confusing them with \textit{CSAI}, \textit{suspected CSAI}, and \textit{porn}. We believe that this confusion is potentially due to the presence of nudity or seminudity without sexual connotations. Considering only \textit{CSAI} and \textit{suspected CSAI} classes and mapping all the others to \textit{not CSAI}, the balanced accuracy increases to \text{53.44 ± 0.16\%}.

We also performed evaluation on the region-based annotated child pornography dataset (RCPD)~\cite{Macedo2018BenchmarkMethodologyChild}, a CSAI benchmark composed of over 2,000 samples among regular and CSAI images, produced by Brazilian Federal Police. We performed only the FSL evaluation protocol, since the dataset does not contain indoor annotation. In each of the 10,000 episodes, five samples per class (10 samples total) were randomly selected for the support set and 15 samples per class (30 samples total) for the query set. 

\begin{figure}[t!]
\centering
\includegraphics[width=\linewidth]{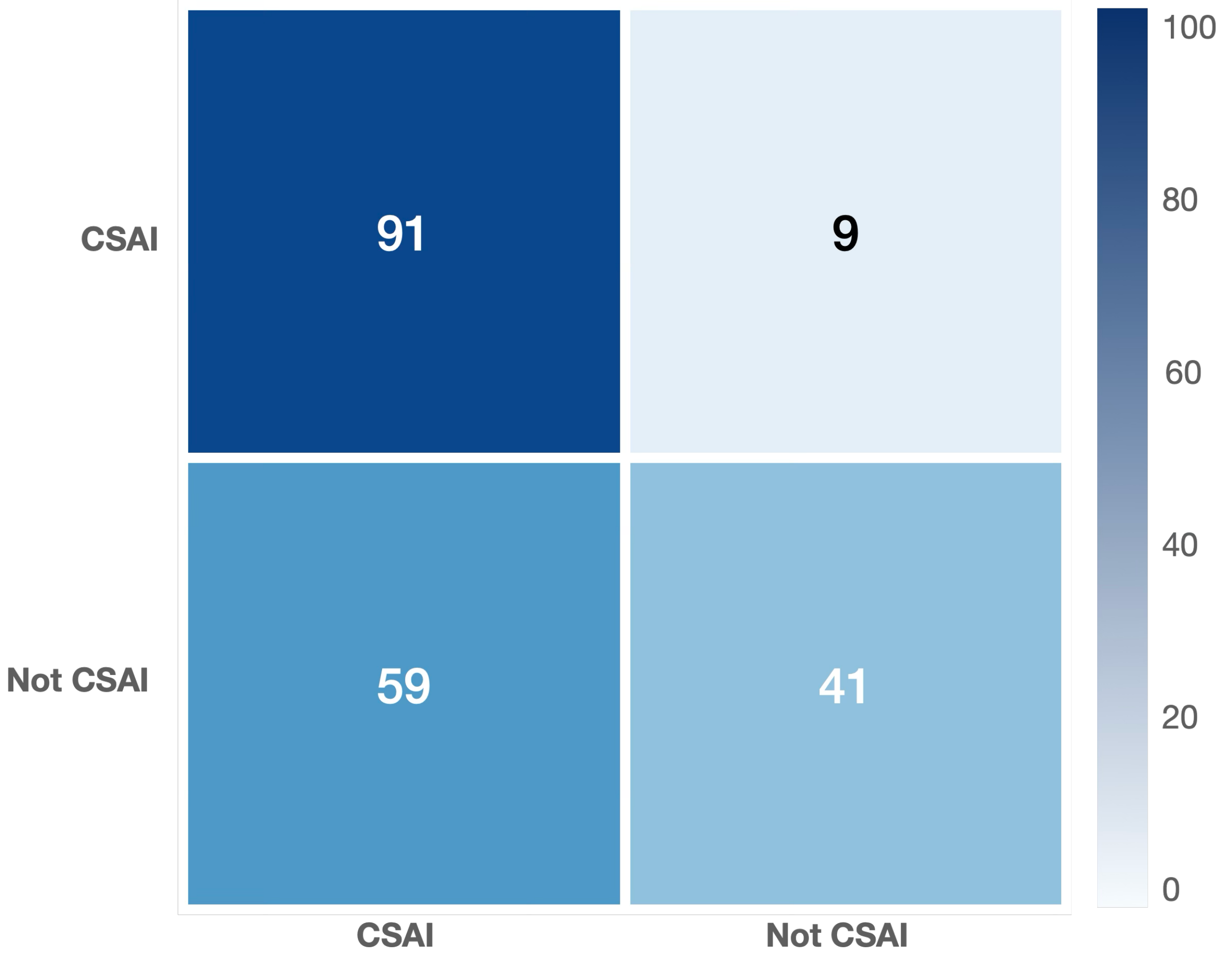}
\caption[Confusion matrix with the model's results on the RCPD dataset.]{Confusion matrix with the model's results on RCPD~\cite{Macedo2018BenchmarkMethodologyChild}. Values reported are the accuracy (\%) for prediction per class. Ground truth labels are on the vertical axis; predicted labels on the horizontal axis.}
\Description{A confusion matrix grid visualization. The grid shows ground truth labels on the vertical axis and predicted labels on the horizontal axis. Each cell is shaded to indicate accuracy or frequency, with darker shades representing higher values. The matrix highlights correct predictions along the diagonal and errors off-diagonal.}
\label{fig:cm_rcpd} 
\end{figure}

The model achieved an average accuracy with 95\% confidence of 65.40 ± 0.16\%. Fig.~\ref{fig:cm_rcpd} shows that the model was good at classifying CSAI samples, while it struggled with non-CSAI samples. This behavior is probably due to images containing nudity or seminudity that were not CSAI. We can see nonetheless that such a model would be useful to immediately identify images that are suspect of being~CSAI.

The results from both datasets indicate that the model can classify samples reasonably well even with little data in the support set. This suggests the potential of using indoor scene classification to help in CSAI investigations. Combined with other complementary features, it could build a robust CSAI~classifier.

\subsection{Case Study Main Takeaways and Future Work}

With this case study we have shown the potential not only of our protocol for investigating future Proxy tasks for CSAI recognition but we have also shown how few-shot learning and indoor scene classification can benefit CSAI investigation. Our results demonstrate the benefits of leveraging pre-trained models and adapting them with limited indoor scene data relevant to CSAI investigations. 

We highlight how the few-shot protocols allow for models to be used on a different domain than what they were tested on and particularly embedding based FSL allows for this use without training model weights on sensitive data. This allows for such models to be shared across law enforcement agencies without fear that sensitive data will leak through model weights. 

While five samples per class is a standard benchmark for FSL, future work could explore varying sample sizes in the support set could optimize performance for CSAI recognition tasks. This investigation could determine the ideal number of samples to balance accuracy and efficiency in real-world applications.

Furthermore, investigating diverse classifiers beyond cosine simi\-larity-based ones, such as Euclidean distance, linear classifiers, or appended neural networks, could bring more accurate results~\cite{sung2018relation, oreshkin2018tadam, mishra2017snail, garcia2017gnn, ssformers, doersch2020crosstransformers}. Further addressing the inherent domain shift between standard datasets and CSAI is also paramount. 

Finally, we would like to raise awareness to a limitation present in our case study and also common to most models trained on Proxy Tasks and used on CSAI. When making use of public benchmarks and public datasets from related tasks we are risking ``contaminating'' the CSAI recognition task with a non-neutral representation of the related sub-task \cite{Koch2021ReducedReusedRecycled} carried forward from the original dataset's intent and biases; this distancing from original purpose to final evaluation further muddles the implications of these data choices. For example, in preliminary talks with LEAs regarding the Places365 dataset they have already pointed out how the depiction of ``bedroom'' and ``child's bedroom'' in Places365 often presents a skewed ``perfect'' or ``magazine-like'' idea of these classes. This bias is certain to have an impact when models are used to identify bedrooms from distinct backgrounds and under more difficult lighting conditions.

\section{Conclusion}

Building a method that will be applied to a dataset that is never seen is a considerable challenge; not only that, but creating a classifier that will be used for data that sometimes does not have a precise classification is also difficult. With this manuscript we wanted to create a path to make more research on CSAI possible. Through a formal definition of Proxy Tasks, authors can have a better chance of comparing their results on public benchmarks for these related sub-tasks of CSAI Recognition. 

We furthermore encourage that these Proxy Tasks are well thought out, that law-enforcement is never left out of the conversation and that we do not forget that both the data and the law-enforcement agencies that store that data will have constraints on how much labeled data there is, how it can be used and how models should or should not interact with such data.

We introduced a common protocol for investigating these Proxy Tasks and showed a case study on few-shot learning for indoor scene classification; through our case study we were able to observe both the benefits and the shortcomings of this Proxy Task. Our expectation is that systems used by LEAs in the future are actually constituted of multiple Proxy Task-based models, each highlighting one aspect of a full investigation of these images, from scene classification to nudity detection and age estimation. 

\section{Ethical Considerations Statement}

All CSAI data was only ever handled by LEAs authorized to do so and CSAI data was never taken out of the premises of partner law enforcement agencies authorized to handle it; we have provided LEAs with our own code for evaluation on their authorized hardware. Furthermore, the goal of this manuscript is to introduce a way to study the problem of child sexual abuse imagery (CSAI) detection while minimizing the interactions between models and CSAI and avoiding entirely interactions between CSAI and practitioners that are not LEAs. We are furthermore interested in galvanizing research on CSAI detection so that LEAs are less likely to have to check CSAI manually, a task that negatively impacts them daily.

While working with partnership LEAs, we prioritized minimizing their interaction with CSAI caused by our research; of note is that our case study experiments with CSAI were only executed once, with the model found to be best suited for the Proxy Task. We hope this practice becomes a pattern in the field, again to avoid increasing the burden on LEAs.

\section{Adverse Impacts Statement}

We list below our considerations of two potential adverse impacts of our work. 

\paragraph{Evaluation only with Proxy Tasks} While we discourage this practice, it is possible that some practitioners that do not have an established partnership with LEAs make use of Proxy Tasks and their attached benchmarks isolated from evaluation on CSAI or conversations with LEAs. While achieving better metrics on Proxy Tasks is encouraged, we highlight the importance of keeping the stakeholders of the final task in mind and always seek their guidance and help with executing final evaluations on real CSAI.

\paragraph{Proxy Tasks and their models could be collected with malicious intent} We have highlighted that one of the advantages of experimenting first with Proxy Tasks is that, because they were not trained and do not carry any information about CSAI within their weights, these models can be freely shared and reproduced, improving potential for collaboration. A malicious actor could collect these models and what aspects of CSAI each model represents to either check their own illegal CSAI against them or use them to search for CSAI in large public datasets.

\section*{Acknowledgments}

This work is partially funded by FAPESP 2023/12086-9, PIND/ FAEPEX UNICAMP 2597/23, and the Serrapilheira Institute R-2011-37776. T.~Coelho is also funded by Becas Santander and Alumni Grant (Instituto de Computação). L.~S.~F.~Ribeiro is also funded by FAPESP 2022/14690-8, S.~Avila is also funded by FAPESP 2020/09838-0, 2013/08293-7, H.IAAC 01245.003479/2024-10, and CNPq 316489/ 2023-9.

\bibliographystyle{ACM-Reference-Format}
\bibliography{bibliography}

\end{document}